\documentclass[letterpaper,11pt]{article}
\usepackage{paper}
\usepackage{float}
\usepackage{algorithm}
\usepackage{algorithmic}

\hypersetup{pdftitle={Dynamics, stability, and energy efficiency of an energy-recycling rimless wheel with spring-clutch legs}}

\newcommand{\bib}{paper.bib}

\begin{document}

\title{Dynamics, stability, and energy efficiency of an energy-recycling rimless wheel with spring-clutch legs}

\author{Tongchen Lin$^\dagger$, Yanqiu Zheng$^\dagger$, Chuhan Zhang, Ruigang Chen, Yizhar Or, Mingyi Liu%
\thanks{Tongchen Lin, Chuhan Zhang, Ruigang Chen, and Mingyi Liu: Department of Mechanical Engineering and Robotics, Guangdong Technion-Israel Institute of Technology, Shantou, China. Tongchen Lin, Chuhan Zhang, Ruigang Chen, Yizhar Or, and Mingyi Liu: Faculty of Mechanical Engineering, Technion-Israel Institute of Technology, Haifa, Israel. Yanqiu Zheng: Department of Applied Electronics, Faculty of Advanced Engineering, Tokyo University of Science, Tokyo, Japan. $^\dagger$These authors contributed equally to this work. Corresponding author: Mingyi Liu, mingyi.liu@gtiit.edu.cn.}}

\date{June 2026}


\begin{titlepage}
\maketitle

This paper proposes an energy-recycling rimless wheel with spring-clutch legs. The proposed mechanism uses a lockable clutch to store part of the impact-induced elastic energy after foot contact and reinject it in the next gait cycle. First, we develop a hybrid dynamic model of the energy-recycling rimless wheel. Second, numerical simulations are used to examine the dynamics, local stability of periodic gaits, and the Cost of Transport (CoT) of the proposed mechanism. The simulation results show that the proposed mechanism reduces the CoT by up to $16.13\%$ compared with a benchmark viscoelastic-legged rimless wheel with telescopic spring-damper legs. Compared with the rigid rimless wheel, the viscoelastic-legged and energy-recycling models reduce the CoT by more than $50\%$. The energy-recycling model also maintains locally stable periodic gaits over the tested slope and stiffness ranges. Finally, prototype experiments on an inclined plane are conducted to examine the feasibility of the proposed mechanism. The experimental results show that the proposed rimless wheel achieves passive walking on a shallow $1^\circ$ slope, corresponding to a CoT of approximately $0.02$. These results suggest that the proposed spring-clutch mechanism can improve the simulated walking efficiency of the energy-recycling rimless wheel, while the prototype experiments support the feasibility of passive walking with the mechanism.

\end{titlepage}

\section{Introduction}
\label{sec:introduction}

Achieving energy-efficient walking is one of the challenging tasks in legged robotics. For rigid-legged robots, mechanical energy is commonly dissipated during foot--ground collisions. Conventional bipedal robots usually compensate for this lost energy through active actuation, which increases mechanical complexity, control difficulty, and energy consumption. In contrast, passive dynamic walking provides a simple example of energy-efficient locomotion using natural dynamics and gravitational input. A passive dynamic walker can walk down a gentle slope without motors or controllers because gravity provides the energy needed to compensate for the mechanical energy dissipated at foot impact \cite{collins2005efficient}. This concept was initially introduced by McGeer \cite{mcgeer1990passive}, and has inspired various energy-efficient walking machines, including quasi-dynamic walkers \cite{1570404, bhounsule2014low, 7270326,9099094}.

As the simplest passive dynamic walker, the rimless wheel has been widely used to study the basic mechanics and energy cost of legged locomotion \cite{coleman2010dynamics, zheng2025tensegrity}. Based on this model, several studies have introduced active control strategies to enable level-ground locomotion while maintaining low energy consumption \cite{bhounsule2016dead, sanchez2021design, hanazawa2018development}. However, energy loss at foot impact remains an inherent limitation of the rigid rimless wheel. During stance-leg exchange, mechanical energy is dissipated through the collision between the swing leg and the ground.

To reduce this impact loss, previous work has explored mechanisms for collisionless walking. Gomes and Ruina \cite{gomes2011walking} proposed a walking model that avoids collision loss by swinging the upper body. This result shows the theoretical possibility of collisionless walking, although no physical prototype was reported. Gomes and Ahlin \cite{gomes2015quiet} later developed a rimless wheel-based prototype that achieves nearly collisionless walking by matching the foot velocity with the ground velocity at impact.

Besides direct impact reduction, introducing leg compliance provides another way to reduce collision-induced energy loss. The main idea is to allow impact energy to be temporarily stored in elastic elements rather than dissipating it immediately at foot contact. This principle is consistent with animal locomotion, where tendons and ligaments act as passive springs that store energy during ground contact and release it during forward motion \cite{alexander1990three}. In robotic locomotion, compliant legs and elastic actuators have been studied for impact reduction, disturbance response, and energy-efficient cyclic motion \cite{pal2026energy, rummel2010stable, hurst2008role}. Properly designed elastic elements can shape the natural dynamics of the robot and reduce the power or energy required under specific gait conditions.

Passive compliance alone cannot control when the stored energy is released. The spring may return this energy at a phase where it does not contribute effectively to the gait cycle. Energy-recycling mechanisms provide a way to regulate this process by temporarily storing mechanical energy and releasing it at a suitable phase of the motion cycle. This concept has been applied to wearable assistive devices, where energy can be stored during heel strike and released to assist push-off \cite{collins2015reducing, collins2010recycling, wang2021design, lee2023lower}. Similar ideas have also been explored in robotic actuation. Krimsky and Collins \cite{krimsky2024elastic} developed an energy-recycling actuator using parallel springs and electroadhesive clutches, achieving up to $97\%$ energy savings across diverse cyclic tasks.

In legged locomotion, BirdBot provides a representative example of clutch-like mechanical coupling for coordinating leg loading and phase transitions. Its avian-inspired multiarticular spring--tendon network engages upon foot contact to form a load-bearing stance leg and disengages near the end of stance to allow low-resistance leg flexion for swing \cite{badri2022birdbot}. These studies suggest that, beyond adding elastic elements, regulating the timing of mechanical engagement and energy release is important for reducing actuation effort and improving the efficiency of cyclic locomotion.

Based on this observation, this paper proposes an energy-recycling rimless wheel with spring-clutch legs. The proposed walker is based on a rimless wheel with telescopic compliant legs, where each leg contains a spring and a lockable clutch. The spring does not release the stored energy immediately after compression, but can be locked by the clutch and released in the subsequent gait cycle. In this way, the mechanism stores impact-related energy through collision-induced leg compression and returns the stored energy at a later phase of the walking cycle.

This work builds upon the viscoelastic-legged rimless wheel model proposed by Asano \cite{asano2012passive}, which clarified the role of double-limb support in passive walking with telescopic compliant legs. Unlike that model, the proposed model adds a lockable clutch to the compliant leg so that energy storage and release can be treated as separate phases. This modification retains the simple rimless wheel structure and passive walking framework while enabling controlled energy release in the subsequent gait cycle.

The main objective of this study is to examine how spring-clutch energy recycling affects passive rimless wheel walking. First, we derive a hybrid dynamic model that includes the single-support phase, double-support phase, clutch constraint, and collision condition for stance-leg exchange. Second, we compute periodic passive walking gaits and evaluate local stability using a Poincaré return map. Third, we compare the Cost of Transport of the proposed mechanism with that of rigid and viscoelastic-legged rimless wheels. Finally, we conduct prototype experiments on an inclined plane to examine the feasibility of passive walking with the proposed mechanism.

The subsequent sections are organized as follows. Section~\ref{sec:modeling} introduces the energy-recycling rimless wheel model and its hybrid dynamics. Section~\ref{sec:numerical-simulation} presents the numerical gait analysis, local stability evaluation, and comparison of the Cost of Transport. Section~\ref{sec:experiments} describes the prototype design and inclined-plane walking tests. Section~\ref{sec:conclusion} summarizes the main findings and discusses the feasibility and limitations of the proposed mechanism.

\section{Modeling}
\label{sec:modeling}

\begin{figure}[H]
      \centering
      \includegraphics{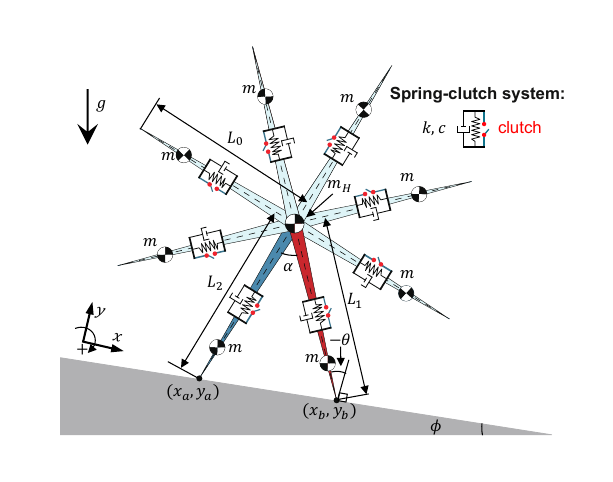}
      \caption{Dynamic model of the energy-recycling rimless wheel}
      \label{model}
\end{figure}

The energy-recycling rimless wheel model is shown in Fig.~\ref{model}. The following constant parameters are used to describe this model. $m_H$ denotes the mass of the main body, and $m$ represents the mass of each leg. $I_c$ denotes the rotational inertia of the energy-recycling rimless wheel about its center. Linear springs and dampers are mounted inside each leg, with spring stiffness $k$ and damping coefficient $c$. $L_0$ is the leg length when the spring is in its uncompressed state. $\alpha$ is the angle between two spokes, which is $\frac{\pi}{4}\,[\mathrm{rad}]$ in this design. $\phi$ denotes the slope angle.

The time-varying quantities of the model are defined as follows. $L_i$ represents the leg length, where $i=1$ indicates the leading leg and $i=2$ indicates the trailing leg. $\theta$ represents the angle between the leading leg and the slope normal. The contact points of the trailing and leading legs are denoted by $(x_a,y_a)$ and $(x_b,y_b)$, respectively. In this model, the displacement of the moving parts in the spring legs is assumed to be negligible; therefore, the moment of inertia of the wheel remains constant. In addition, no-slip contact with the ground is assumed.

\begin{figure}[H]
      \centering
      \includegraphics{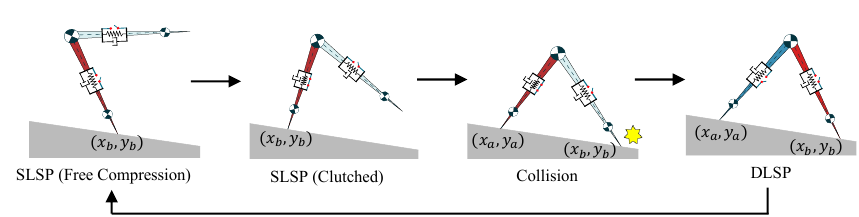}
      \caption{State transitions of the energy-recycling rimless wheel model between three states (SLSP (Free Compression), SLSP (Clutched), DLSP).}
      \label{state}
\end{figure}

The state transitions of the energy-recycling rimless wheel are summarized in Fig.~\ref{state}. The system has two motion phases: the single-legged support phase (SLSP) and the double-legged support phase (DLSP). The SLSP includes two stance-leg states. In SLSP (Free Compression), the stance-leg spring is compressed until clutch engagement. In SLSP (Clutched), the stance-leg spring is locked by the clutch until the swing leg collides with the ground. At this collision, the stance-leg clutch disengages, and the system transitions to the DLSP.

\subsection{Equation of Motion}

For the three constrained states, SLSP (Free Compression), SLSP (Clutched), and DLSP, the equations of motion are given as follows.

\subsubsection{Single-legged Support Phase (Free Compression)}

Let $\mathbf{q} = \begin{bmatrix} x_b & y_b & \theta & L_1 & L_2 \end{bmatrix}^{\mathrm{T}}$ be the generalized coordinate vector. During SLSP (Free Compression), the equation of motion is written as:
\begin{equation}
\mathbf{M}(\mathbf{q})\ddot{\mathbf{q}} + \mathbf{h}(\mathbf{q},\dot{\mathbf{q}}) = \mathbf{J}_b^{\mathrm{T}}\boldsymbol{\lambda}_b,
\label{1}
\end{equation}
where $\mathbf{M}$ is the inertia matrix, and $\mathbf{h}$ represents the combined centrifugal, Coriolis, gravitational, spring, and damping terms. $\mathbf{J}_b$ denotes the Jacobian matrix associated with the holonomic constraints at the grounding point $(x_b,y_b)$, and $\boldsymbol{\lambda}_b$ is the constraint force vector.

The grounding point of the leading leg is fixed based on the assumption of no slipping; the constraint equation is expressed as follows:

\begin{equation}
    \dot{x}_b=0, \quad \dot{y}_b=0.
    \label{2}
\end{equation}

Accordingly, the constraint Jacobian matrix $\mathbf{J}_b$ is obtained from Eq.~(\ref{2}) as:

\begin{equation}
\mathbf{J}_b \dot{\mathbf{q}} = 
\begin{bmatrix}
1 & 0 & 0 & 0 & 0\\
0 & 1 & 0 & 0 & 0
\end{bmatrix}
\dot{\mathbf{q}} = \mathbf{0}_{2 \times 1}.
\label{3}
\end{equation}

By substituting the time derivative of Eq.~(\ref{3}) into Eq.~(\ref{1}), the constraint force vector $\boldsymbol{\lambda}_b$ can be obtained:

\begin{equation}
\boldsymbol{\lambda}_b
=
\begin{bmatrix}
\lambda_{b,x} \\
\lambda_{b,y}
\end{bmatrix}
=
\left( \mathbf{J}_b \mathbf{M}^{-1} \mathbf{J}_b^{\mathrm{T}} \right)^{-1}
\left( \mathbf{J}_b \mathbf{M}^{-1} \mathbf{h} \right),
\label{4}
\end{equation}
where $\lambda_{b,x}$ and $\lambda_{b,y}$ represent the horizontal and vertical ground reaction forces at the grounding point $(x_b,y_b)$, respectively.

\subsubsection{Single-legged Support Phase (Clutched)}

When the stance-leg spring reaches its maximum compression, the clutch engages and locks the spring. As a result, the stance-leg length $L_1$ is constrained to remain constant. In this state, the system is subject to both the ground constraint at the stance foot and the internal clutch constraint. The equation of motion becomes:

\begin{equation}
\mathbf{M}\ddot{\mathbf{q}} + \mathbf{h}
= \mathbf{J}_{cl}^{\mathrm{T}}\boldsymbol{\lambda}_{cl},
\label{2.5}
\end{equation}
where $\mathbf{J}_{cl}$ is the Jacobian matrix associated with the holonomic constraints at the grounding point $(x_b,y_b)$ and the clutch constraint on $L_1$, and $\boldsymbol{\lambda}_{cl}$ is the constraint force vector.

The corresponding constraint equations are expressed as follows:
\begin{equation}
    \dot{x}_b=0, \quad \dot{y}_b=0, \quad \dot{L}_1=0.
\end{equation}

Accordingly, the constraint Jacobian matrix $\mathbf{J}_{cl}$ can be summarized as:
\begin{equation}
\mathbf{J}_{cl} \dot{\mathbf{q}} = 
\begin{bmatrix}
1 & 0 & 0 & 0 & 0\\
0 & 1 & 0 & 0 & 0\\
0 & 0 & 0 & 1 & 0
\end{bmatrix}
\dot{\mathbf{q}} = \mathbf{0}_{3 \times 1}.
\label{2.7}
\end{equation}

Solving Eq.~(\ref{2.5}) together with the time derivative of Eq.~(\ref{2.7}) gives the constraint force vector $\boldsymbol{\lambda}_{cl}$ as:
\begin{equation}
\boldsymbol{\lambda}_{cl}
=
\begin{bmatrix}
\lambda_{b,x} \\
\lambda_{b,y} \\
\lambda_{L_1}
\end{bmatrix}=
\left( \mathbf{J}_{cl} \mathbf{M}^{-1} \mathbf{J}_{cl}^{\mathrm{T}} \right)^{-1}
\left( \mathbf{J}_{cl} \mathbf{M}^{-1} \mathbf{h} \right),
\end{equation}
where $\lambda_{L_1}$ is the clutch constraint force that maintains $L_1$ constant after clutch engagement.

The transition from SLSP to DLSP is triggered when the swing leg contacts the ground. The corresponding event condition is defined as follows:

\begin{equation}
\begin{aligned}
&L_1\cos(\theta+\phi)
-L_0\cos(\alpha-\theta-\phi) \\
&+\sin\phi\sqrt{L_1^2+L_0^2-2L_1L_0\cos\alpha}
=0.
\end{aligned}
\end{equation}
This condition represents the geometric contact condition between the swing leg and the ground surface.

\subsubsection{Double-legged Support Phase}

In this phase, both feet are constrained by the ground, and both leg lengths are allowed to vary because the clutches are disengaged. The equation of motion is expressed as follows:

\begin{equation}
\mathbf{M} \ddot{\mathbf{q}} + \mathbf{h} = \mathbf{J}_{ab}^{\mathrm{T}} \boldsymbol{\lambda}_{ab},
\end{equation}
where $\mathbf{J}_{ab}$ represents the Jacobian matrix for the holonomic constraints at the grounding points $(x_a, y_a)$ and $(x_b, y_b)$, and $\boldsymbol{\lambda}_{ab}$ is the constraint force vector.

The velocity constraints are written as follows:

\begin{equation}
\dot{x}_b = 0, \quad \dot{y}_b = 0, \quad \dot{x}_a = 0, \quad \dot{y}_a = 0.
\end{equation}

Hence, the constraint Jacobian matrix $\mathbf{J}_{ab}$ is written as:
\begin{equation}
\mathbf{J}_{ab} \dot{\mathbf{q}} = 
\begin{bmatrix}
1 & 0 & 0 & 0 & 0 \\
0 & 1 & 0 & 0 & 0 \\
1 & 0 & J_{33} & \sin\theta & J_{35} \\
0 & 1 & J_{43} & \cos\theta & J_{45}
\end{bmatrix}\dot{\mathbf{q}}=\mathbf{0}_{4 \times 1},
\label{Jacobian}
\end{equation}
where
\begin{equation}
\begin{aligned}
J_{33} &= L_1 \cos\theta - L_2 \cos(\theta + \alpha), \quad &&J_{35} = -\sin(\theta + \alpha), \\
J_{43} &= -L_1 \sin\theta + L_2 \sin(\theta + \alpha), \quad &&J_{45} = -\cos(\theta + \alpha).
\end{aligned}
\end{equation}

Following the derivation of $\boldsymbol{\lambda}_b$ in Eq.~(\ref{4}), the constraint force vector $\boldsymbol{\lambda}_{ab}$ is obtained as follows:

\begin{equation}
    \boldsymbol{\lambda}_{ab} = \begin{bmatrix}
\lambda_{b,x} \\
\lambda_{b,y} \\
\lambda_{a,x} \\
\lambda_{a,y}
\end{bmatrix}=\left( \mathbf{J}_{ab} \mathbf{M}^{-1} \mathbf{J}_{ab}^{\mathrm{T}} \right)^{-1}
\left( \mathbf{J}_{ab} \mathbf{M}^{-1} \mathbf{h} - \dot{\mathbf{J}}_{ab} \dot{\mathbf{q}} \right),
\end{equation}
where $\lambda_{a,x}$ and $\lambda_{a,y}$ represent the horizontal and vertical ground reaction forces at the grounding point $(x_a,y_a)$, respectively.

Here, the transition from DLSP to SLSP is triggered when the trailing leg's vertical ground reaction force reaches zero while the leading leg maintains contact with the ground, i.e., $\lambda_{a,y}=0$ and $\lambda_{b,y}>0$.

\subsubsection{Collision Equation}

At the instant of collision, the leg labels are exchanged so that the new stance leg is described by the same generalized coordinates in the next phase. A leg-exchange mapping is applied to the
pre-impact states $\mathbf{q}^{-}$ and $\dot{\mathbf{q}}^{-}$, yielding the
relabeled states $\tilde{\mathbf{q}}^{-}$ and $\dot{\tilde{\mathbf{q}}}^{-}$ as follows:

\begin{equation}
\tilde{\mathbf{q}}^{-} =
\begin{bmatrix}
x^{-} +L_1^{-}\sin{\theta^{-}} + L_0\sin{(\alpha-\theta^{-})}\\
y^{-} +L_1^{-}\cos{\theta^{-}} - L_0\cos{(\alpha-\theta^{-})} \\
\theta^{-} - \alpha \\
L_0 \\
L_1^{-}
\end{bmatrix},
\qquad
\dot{\tilde{\mathbf{q}}}^{-} =
\begin{bmatrix}
\dot{\tilde{x}}^{-} \\
\dot{\tilde{y}}^{-} \\
\dot{\theta}^{-} \\
0 \\
\dot{L}_1^{-}
\end{bmatrix},
\end{equation}

\begin{equation}
\dot{\tilde{x}}^{-}
=
\dot{x}^{-}
+ \left( L_1^{-}\cos\theta^{-} - L_0\cos(\alpha - \theta^{-}) \right)\dot{\theta}^{-}
+ \dot{L}_1^{-}\sin\theta^{-},
\end{equation}

\begin{equation}
\dot{\tilde{y}}^{-}
=
\dot{y}^{-}
- \left( L_1^{-}\sin\theta^{-} + L_0\sin(\alpha - \theta^{-}) \right)\dot{\theta}^{-}
+ \dot{L}_1^{-}\cos\theta^{-}.
\end{equation}

Before impact, the system remains in the SLSP (Clutched), where the clutch of the stance leg constrains $L_1$. After the swing leg collides with the ground, this clutch is assumed to disengage immediately, and the system transitions to the DLSP. Therefore, both leg springs are free to move after impact, and the clutch constraint on $L_1$ is not included in the impact constraint.

Assuming that the ground is rigid and that the collision between the swing leg and the ground is perfectly inelastic, the impact equation is written as follows:

\begin{equation}
\mathbf{M}(\tilde{\mathbf{q}}^{-})\,\dot{\mathbf{q}}^{+}
= \mathbf{M}(\tilde{\mathbf{q}}^{-})\,\dot{\tilde{\mathbf{q}}}^{-}
+ \mathbf{J}_c(\tilde{\mathbf{q}}^{-})^{\mathrm{T}}\,\boldsymbol{\lambda}_c,
\label{collision1}
\end{equation}

\begin{equation}
\mathbf{J}_c(\tilde{\mathbf{q}}^{-})\,\dot{\mathbf{q}}^{+}
= \mathbf{0}_{4 \times 1},
\label{collision2}
\end{equation}
where the superscripts $-$ and $+$ denote the states immediately before and after the collision, respectively. The matrix $\mathbf{J}_c$ is the impact constraint Jacobian, and $\boldsymbol{\lambda}_c$ denotes the impulsive constraint vector generated during collision. The Jacobian matrix $\mathbf{J}_c(\tilde{\mathbf{q}}^{-}) \in \mathbb{R}^{4\times 5}$
is identical to Eq.~(\ref{Jacobian}). By solving Eqs.~(\ref{collision1}) and (\ref{collision2}) simultaneously, $\dot{\mathbf{q}}^{+}$ is obtained as:

\begin{equation}
\dot{\mathbf{q}}^{+}
=
\left(
\mathbf{I}_5
-
\mathbf{M}(\tilde{\mathbf{q}}^{-})^{-1}
\mathbf{J}_c(\tilde{\mathbf{q}}^{-})^{\mathrm{T}}
\left(
\mathbf{J}_c(\tilde{\mathbf{q}}^{-})
\mathbf{M}(\tilde{\mathbf{q}}^{-})^{-1}
\mathbf{J}_c(\tilde{\mathbf{q}}^{-})^{\mathrm{T}}
\right)^{-1}
\mathbf{J}_c(\tilde{\mathbf{q}}^{-})
\right)
\dot{\tilde{\mathbf{q}}}^{-}.
\end{equation}

Since the post-impact motion is assumed to enter the DLSP, both contact points must remain in contact with the ground immediately after impact. Therefore, the normal components of the contact impulses, $\lambda_{b,y}^{c}$ and $\lambda_{a,y}^{c}$, must be positive:

\begin{equation}
\lambda_{b,y}^{c} > 0, \quad \lambda_{a,y}^{c} > 0,
\label{impact_impulse_condition}
\end{equation}
where $\lambda_{b,y}^{c}$ and $\lambda_{a,y}^{c}$ denote the components of the impact impulses along the slope-normal direction at the contact points $(x_b,y_b)$ and $(x_a,y_a)$, respectively. These inequalities ensure that the ground applies compressive, rather than tensile, impulses at both contacts. In the simulations presented in this work, these conditions are checked at every impact to validate the transition to double support.

After the impact calculation, the post-impact state $\mathbf{q}^{+}$ and velocity $\dot{\mathbf{q}}^{+}$ are used as the initial conditions for the subsequent DLSP.

\subsection{Hybrid Periodic Solution and Orbital Stability}

To compare the energy efficiency of systems with and without a clutch on the same basis, their periodic solutions and stability must be analyzed. The existence and stability of periodic solutions in hybrid dynamic systems have been extensively studied in the literature, typically through the use of the Poincaré map \cite{gamus2015dynamic}.

The state vector of the hybrid system is defined as:
\begin{equation}
    \mathbf{x} = (\mathbf{q}, \dot{\mathbf{q}}),
    \label{eq:state_vector}
\end{equation}
where $\mathbf{q}$ and $\dot{\mathbf{q}}$ denote the generalized coordinates and generalized velocities, respectively.

Since the Poincaré map is constructed from one post-impact state to the next, the Poincaré section is chosen as the post-impact section:
\begin{equation}
    \Sigma^{+}
    =
    \left\{
    \mathbf{x}^{+}
    :
    \mathbf{x}^{+}
    =
    \mathbf{x}(t_{\mathrm{c}}^{+})
    \right\},
    \label{eq:poincare_section_general}
\end{equation}
where $t_{\mathrm{c}}^{+}$ denotes the instant immediately after a foot--ground collision. Since $\Sigma^{+}$ is a section of the state space, its dimension is lower than that of the full state space. Therefore, the Poincaré map is evaluated on this lower-dimensional section.

The Poincaré map is then defined as $\Pi:\Sigma^{+}\rightarrow\Sigma^{+}$, which maps an initial post-impact state $\mathbf{x}_0\in\Sigma^{+}$ to the post-impact state immediately after the next collision:
\begin{equation}
    \Pi(\mathbf{x}_0)
    =
    \mathbf{x}(t_{\mathrm{c}}^{+}).
    \label{eq:poincare_map_general}
\end{equation}

The Poincaré map induces a discrete-time dynamical system that governs the step-to-step transition of a hybrid solution:
\begin{equation}
    \mathbf{x}_{k+1} = \Pi(\mathbf{x}_k),
    \label{eq:poincare_discrete_map}
\end{equation}
where \(\mathbf{x}_k = \mathbf{x}(t_k^+)\) denotes the sequence of post-impact states.

A fixed point \(\mathbf{x}^*\) of the Poincaré map satisfies \(\Pi(\mathbf{x}^*) = \mathbf{x}^*\), and corresponds to the initial condition of a periodic solution of the hybrid system. For a passive dynamic walking system, such a fixed point yields a cyclic gait. The periodic solution is said to be orbitally asymptotically stable if \(\mathbf{x}^*\) is asymptotically stable under the discrete-time dynamics in Eq.~\eqref{eq:poincare_discrete_map}.

According to classical stability theory, the stability of \(\mathbf{x}^*\) is determined by the eigenvalues of the linearization matrix of the Poincaré map on $\Sigma^{+}$:

\begin{equation}
    \mathbf{A}
    =
    \left.
    \frac{d\Pi}{d\mathbf{x}}
    \right|_{\mathbf{x}=\mathbf{x}^{*}}.
    \label{eq:poincare_linearization}
\end{equation}
Here, the derivative is understood with respect to a local parametrization of the Poincaré section $\Sigma^{+}$.

The fixed point \(\mathbf{x}^*\) is asymptotically stable if and only if all eigenvalues \(\lambda_i\) of \(\mathbf{A}\) lie strictly within the unit circle in the complex plane, i.e., \(|\lambda_i| < 1\) for all \(i\).

\subsection{Cost of Transport}

In this subsection, a rigid rimless wheel is used as a reference model for energy-efficiency analysis. The rigid model is obtained by locking all legs of the energy-recycling rimless wheel, and the parameters are identical to those shown in Fig.~\ref{model}.

To quantify the energy efficiency of passive downhill walking, the Cost of Transport (CoT) is defined as

\begin{equation}
\mathrm{CoT}
=
\frac{E}{M g s_x},
\end{equation}
where $E$, $s_x$, and $Mg$ represent the consumed energy, horizontal displacement, and weight of the rimless wheel, respectively. In passive downhill walking, the consumed energy is supplied by the decrease in gravitational potential energy, given by $E = M g s_z$, where $s_z$ is the vertical displacement. Therefore,

\begin{equation}
\mathrm{CoT}
=
\tan\phi
\approx \phi.
\end{equation}

Under the small-slope assumption, the CoT of passive downhill walking can therefore be approximated by the slope angle. Thus, the minimum CoT is defined by the critical slope angle at which a stable periodic gait can be maintained.

For a rigid rimless wheel, Ref.~\cite{coleman2010dynamics} gives the expression for calculating the critical slope angle:

\begin{equation}
g(\phi)=1 - \cos \phi \cos\left(\frac{\alpha}{2}\right) - \left( \frac{1 + \beta^2}{1 - \beta^2} \right) \sin \phi \sin\left(\frac{\alpha}{2}\right) = 0,
\label{rigid angle}
\end{equation}
where

\begin{equation}
\beta = \frac{I_c + ML_0^2 \cos \alpha}{I_c + ML_0^2}.
\end{equation}

The root of $g(\phi) = 0$ represents the minimum critical slope angle $\phi_c$. For any slope where $\phi < \phi_c$, the rimless wheel will fail to cross the potential energy barrier and cannot maintain a forward gait, even if a mathematical fixed point exists. Moreover, the normal ground reaction force must remain positive throughout the step to ensure physical validity.

Since the proposed model is nonlinear, the critical slope angle is computed numerically in the next section, and the resulting CoT is compared with that of the rigid rimless wheel.

\section{Numerical Simulation}
\label{sec:numerical-simulation}

This section compares the viscoelastic-legged and energy-recycling rimless wheels through numerical simulations. The comparison focuses on gait characteristics, periodic solutions and their local stability, and energy efficiency estimated from the minimum stable slope. The gait simulations are calculated using the parameter values listed in Table~\ref{table1}. 

\begin{table}[H]
\caption{Physical parameter settings\label{table1}}
\centering
\begin{tabular*}{0.6\textwidth}{@{\extracolsep{\fill}}lll@{}}
\noalign{\hrule height .6pt}
\textbf{Symbol} & \textbf{Value} & \textbf{Unit}\\
\noalign{\vskip3pt\hrule height .15pt\vskip3pt}
$m_H$ & 1.892 & kg\\
$m$ & 0.0378 & kg\\
$I_c$ & 0.036 & $\mathrm{kg \cdot m^2}$ \\
$L_0$ & 0.237 & m\\
$k$ & 2700 & N/m\\
$c$ & 8 & kg/s\\
$\phi$ & 0.05 & rad\\
$g$ & 9.81 & $\mathrm{m/s^2}$ \\
\noalign{\vskip3pt\hrule height .15pt}
\end{tabular*}
\end{table}

\subsection{Typical Gaits}

\begin{figure}[H]
      \centering
      \includegraphics[width=1\textwidth]{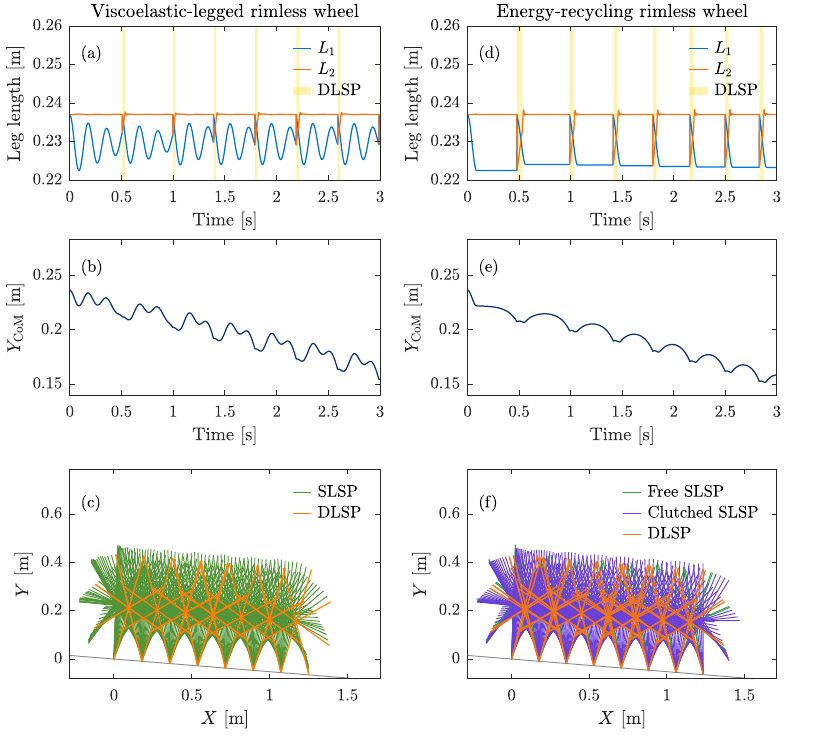}
      \caption{Typical gait trajectories of the viscoelastic-legged and energy-recycling rimless wheels}
      \label{fig:typical_gaits}
\end{figure}

Figure~\ref{fig:typical_gaits} compares the typical passive downhill gaits of the viscoelastic-legged and energy-recycling rimless wheels.
Both systems exhibit continuous passive walking on the slope.
The leg-length plots in Fig.~\ref{fig:typical_gaits}(a) and (d) show the main difference between the two gaits.
In the viscoelastic-legged model, the spring leg is compressed after foot contact, rebounds immediately, and then shows local oscillations.
The spring energy is therefore released immediately after compression.
Because the leg includes damping, part of this oscillatory energy is dissipated during this process.
In the energy-recycling model, the leg length instead follows a compression--locking--release pattern.
After compression, the clutch keeps the spring compressed for a finite time rather than allowing an immediate rebound.
The stored spring energy is then released back into the system during a later gait phase.
Thus, the clutch separates the compression induced by foot contact from the subsequent release of the stored energy.

The CoM trajectories in Fig.~\ref{fig:typical_gaits}(b) and (e) are consistent with the leg-length responses.
In the viscoelastic-legged model, the repeated rebound of the spring leg appears as local oscillations in the CoM height.
In the energy-recycling model, the clutch limits the free vibration of the spring leg, and the local oscillations in the CoM height are reduced.
Thus, the whole-body motion is less affected by spring-leg vibration in this gait.

The stick diagrams in Fig.~\ref{fig:typical_gaits}(c) and (f) show the phase evolution of the two gaits.
In the energy-recycling gait, the free-compression, clutched, and double-support phases can be clearly identified.
This illustrates how the clutch changes the timing of spring compression and release during walking.

\subsection{Ground Reaction Forces}

\begin{figure}[H]
      \centering
      \includegraphics[width=0.9\textwidth]{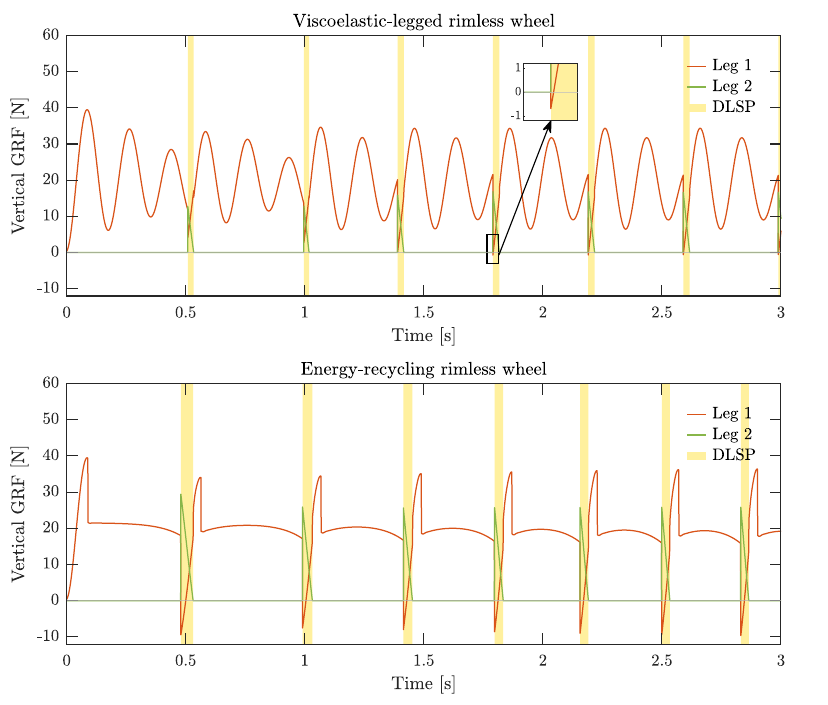}
      \caption{Vertical ground reaction forces of the viscoelastic-legged and energy-recycling rimless wheels during passive downhill walking}
      \label{fig:ground_reaction_forces}
\end{figure}

Figure~\ref{fig:ground_reaction_forces} compares the vertical ground reaction forces of the viscoelastic-legged and energy-recycling rimless wheels during passive downhill walking. This comparison is used to examine whether the rigid-contact formulation produces physically consistent contact forces during the support transition.
In both models, the leading leg shows a short negative vertical force near the beginning of the double-support phase.
This response is nonphysical under unilateral contact, because the ground can only exert a compressive normal force on the foot.

This transient mainly results from the ideal rigid-contact formulation used in the simulation.
At the transition to double support, the holonomic contact constraints at both feet are imposed instantaneously, and the contact forces are obtained as Lagrange multipliers.
This treatment can produce a short tensile-force transient during rapid load transfer between the two legs.
Accordingly, this response should be interpreted as a limitation of the simplified contact model rather than as a specific effect of the spring-clutch mechanism.

Within the present constrained model, increasing the leg damping coefficient $c$ can reduce this transient by suppressing rapid leg-length compression and rebound.
A larger damping coefficient smooths the contact-force response, whereas a smaller damping value allows faster rebound and stronger force oscillation after contact.
The same damping value and contact formulation are used for both models to keep the comparison consistent.
A compliant-ground or penalty-based contact model would provide a more physical treatment of contact establishment and separation by generating normal force only under compressive contact, and will be considered in future work.

\subsection{Energy Transition Ratio Analysis}

\begin{figure}[H]
      \centering
      \includegraphics{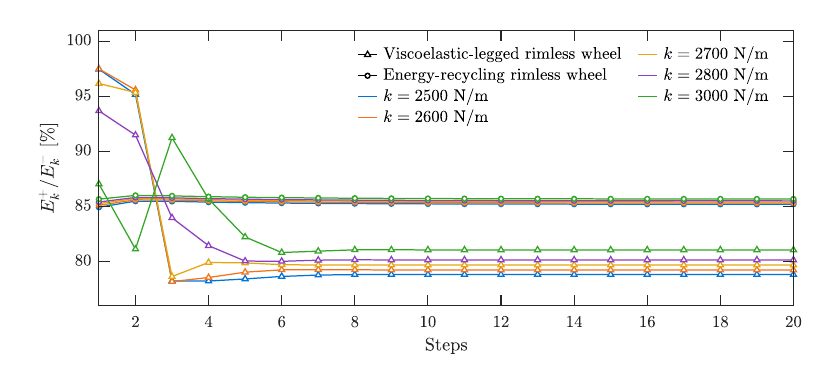}
      \caption{Energy transition ratios at ground collision for the viscoelastic-legged and energy-recycling rimless wheels}
      \label{fig:energy_ratio}
\end{figure}

To evaluate the kinetic-energy loss at ground collision, the energy transition ratio was computed as
$E_k^+ / E_k^-$, where
$E_k^- = \dot{\mathbf{q}}^{-\mathrm{T}} \mathbf{M} \dot{\mathbf{q}}^{-} / 2$
and
$E_k^+ = \dot{\mathbf{q}}^{+\mathrm{T}} \mathbf{M} \dot{\mathbf{q}}^{+} / 2$
represent the kinetic energies before and after collision, respectively.
As shown in Fig.~\ref{fig:energy_ratio}, the transition ratio of the viscoelastic-legged rimless wheel fluctuated during the first few steps and then converged to approximately $79\%$--$81\%$.
The corresponding kinetic-energy loss was about $19\%$--$21\%$ at each ground collision.
In contrast, the energy-recycling rimless wheel maintained a higher transition ratio of approximately $85\%$, corresponding to a kinetic-energy loss of about $15\%$.

The energy-recycling rimless wheel also had a narrower distribution of transition ratios over different spring stiffnesses.
This behavior can be explained by the clutch mechanism. In the viscoelastic-legged model, changing the spring stiffness directly changes the passive compression and rebound process after impact, which leads to different post-impact energy transitions.
In the energy-recycling model, however, the clutch locks the compressed spring and limits this immediate passive rebound.
As a result, the kinetic energy retained after collision becomes less sensitive to the spring stiffness. This may help the system enter the next step with a more consistent energy state within the tested stiffness range.

\subsection{Periodic Solutions and Their Stability}

Algorithm~\ref{alg:critical_slope} summarizes the numerical procedure used to identify the critical slope angle for periodic walking.
Starting from a slope angle at which a periodic gait is found, the slope angle is gradually reduced.
At each slope angle, a fixed point of the Poincaré map is computed using a numerical root-finding method, with the solution obtained at the previous slope used as the initial guess to ensure continuation.

If the fixed-point computation converges, the corresponding slope angle and periodic state are recorded and the procedure proceeds to the next slope.
When convergence fails, the last successfully converged slope angle is identified as the critical slope.
The local stability of the periodic gait at each slope is evaluated by analyzing the eigenvalues of the linearized Poincaré map.
A periodic gait is considered asymptotically stable if the maximum absolute eigenvalue is less than unity.

\begin{algorithm}[H]
\caption{Identification of the Critical Slope Angle via Poincar\'e Analysis}
\label{alg:critical_slope}
\begin{algorithmic}[1]
\REQUIRE Initial slope angle $\phi_{\mathrm{start}}$, step size $\Delta \phi < 0$, minimum slope angle $\phi_{\min}$, initial post-impact state guess $x_0 \in \Sigma^{+}$
\ENSURE Critical slope angle $\phi_c$ and corresponding periodic post-impact state $x_c$

\STATE Set $\phi \leftarrow \phi_{\mathrm{start}}$
\STATE Initialize: $\phi_c \leftarrow \phi$, \ $x_c \leftarrow x_0$

\WHILE{$\phi \ge \phi_{\min}$}
    \STATE Set the global slope angle to $\phi$
    \STATE Solve the fixed-point problem on $\Sigma^{+}$:
    \[
        \Pi(x) - x = 0.
    \]
    \IF{the solver converges successfully}
        \STATE Update initial guess: $x_0 \leftarrow x$
        \STATE Record last converged solution: $\phi_c \leftarrow \phi$, \ $x_c \leftarrow x$
        \STATE Update slope angle: $\phi \leftarrow \phi + \Delta \phi$
    \ELSE
        \STATE \textbf{break}
    \ENDIF
\ENDWHILE

\STATE Compute the Jacobian of the Poincar\'e map at the fixed point:
\[
    \mathbf{A} = \left.\frac{d\Pi}{dx}\right|_{x=x_c}.
\]
\STATE Evaluate the eigenvalues $\{\lambda_i\}$ of $\mathbf{A}$; the periodic gait is
asymptotically stable if $\max_i |\lambda_i| < 1$

\end{algorithmic}
\end{algorithm}

\begin{figure}[H]
      \centering
      \includegraphics{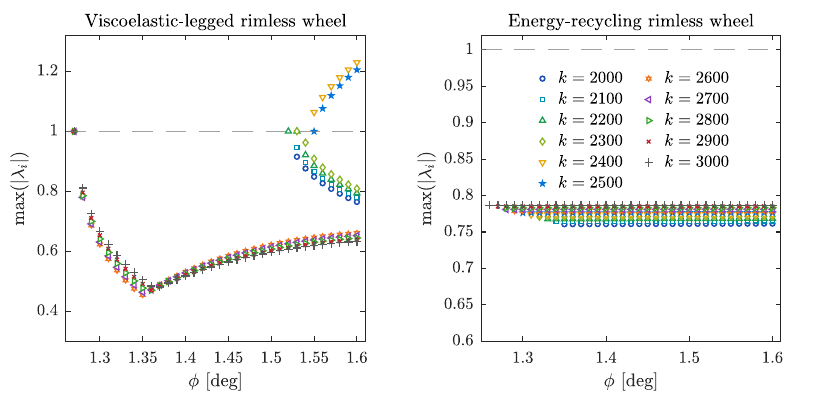}
      \caption{Maximum absolute eigenvalue of the Poincar\'e return map for the viscoelastic-legged rimless wheel and the energy-recycling rimless wheel under different spring stiffness values. The dashed line indicates the unit stability boundary.}
      \label{fig:eigenvalue_comparison}
\end{figure}

Figure~\ref{fig:eigenvalue_comparison} shows the maximum absolute eigenvalue of the Poincaré return map. For the viscoelastic-legged rimless wheel, this stability measure depends strongly on both the slope angle and the spring stiffness.
In the larger-slope region, approximately $\phi=1.55^\circ$--$1.60^\circ$, different stiffness values lead to different stability characteristics. For lower stiffness values, roughly $k=2000$--$2300\,\mathrm{N/m}$, the maximum absolute eigenvalue remains below the unit stability boundary, indicating locally stable periodic gaits. In contrast, for intermediate stiffness values around $k=2400$--$2500\,\mathrm{N/m}$, this value exceeds unity, indicating locally unstable periodic gaits on those branches.

In the intermediate-slope region, approximately $\phi=1.35^\circ$--$1.50^\circ$, several branches remain well below the unit boundary. In particular, some branches reach $\max_i|\lambda_i|\approx0.45$--$0.5$ near $\phi\approx1.35^\circ$, suggesting a relatively large local stability margin. However, near the critical-slope region, around $\phi=1.27^\circ$--$1.30^\circ$, some solutions approach the unit boundary again, indicating a reduced stability margin. The local stability of the viscoelastic-legged model is therefore parameter-sensitive and branch-dependent, rather than being governed by a single monotonic stiffness effect.

In contrast, the energy-recycling rimless wheel maintains a maximum absolute eigenvalue of approximately $0.75$--$0.8$ over the tested slope and stiffness ranges, with all values remaining below unity. The eigenvalue distribution is also much more concentrated than that of the viscoelastic-legged model. The periodic gaits of the energy-recycling rimless wheel therefore remain locally stable over these tested ranges, and the concentrated eigenvalue distribution is consistent with more regular step-to-step dynamics.

\subsection{Energy Efficiency}

\begin{figure}[H]
      \centering
      \includegraphics{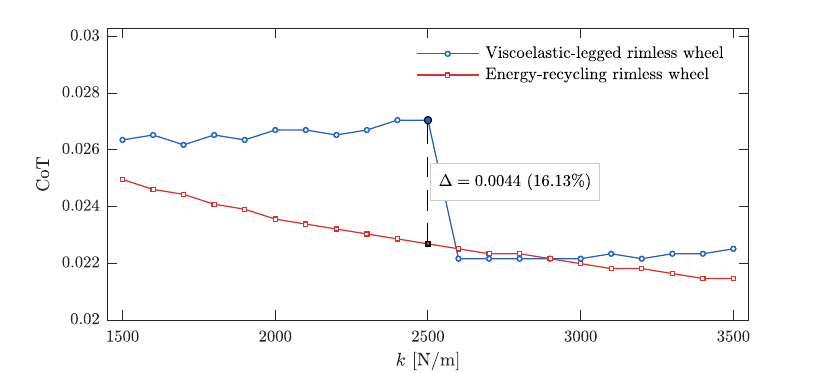}
      \caption{CoT comparison between the viscoelastic-legged and energy-recycling rimless wheels under different spring stiffness values}
      \label{CoT}
\end{figure}

The energy efficiency of the viscoelastic-legged and energy-recycling rimless wheels was evaluated across spring stiffness values from $1500\,\mathrm{N/m}$ to $3500\,\mathrm{N/m}$, as shown in Fig.~\ref{CoT}. For each stiffness value, the critical slope angle was obtained from the Poincar\'e-based continuation procedure described above, and the corresponding CoT was calculated from this angle and used as the efficiency metric. This comparison therefore reflects the minimum slope input required to sustain a locally stable periodic gait, rather than a direct measurement of actuator energy consumption. The energy-recycling rimless wheel exhibits a lower CoT than the viscoelastic-legged model over the tested stiffness range. The maximum reduction is $16.13\%$, obtained at $k = 2500\,\mathrm{N/m}$.

The viscoelastic-legged rimless wheel also shows a lower CoT when the spring stiffness is properly selected. Its CoT decreases sharply near $k = 2600\,\mathrm{N/m}$ and then remains nearly constant for larger stiffness values. Nevertheless, the energy-recycling model maintains a lower CoT in this comparison, indicating that the spring-clutch mechanism can improve the simulated passive walking efficiency without relying on a larger critical slope.

We further compare the two compliant-legged systems with the rigid rimless wheel. The critical slope angle for the rigid rimless wheel is obtained from Eq.~(\ref{rigid angle}). By substituting the parameters listed in Table~\ref{table1}, the numerical result gives a critical slope of $2.80^\circ$, with a corresponding CoT of $0.048\approx0.05$. Both compliant-legged models therefore reduce the CoT by more than $50\%$ compared with the rigid rimless wheel, and the energy-recycling model provides the lowest CoT among the three cases considered here.

\section{Experiments}
\label{sec:experiments}

\label{chap:hardware}

\subsection{Mechanical Design}

\begin{figure}[H]
 \centering
 \includegraphics{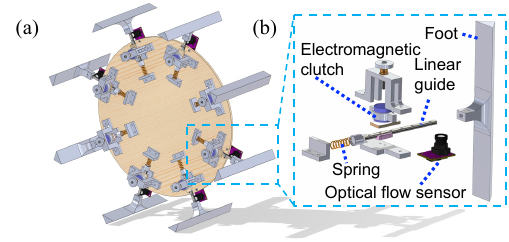}
 \caption{Mechanical design of the energy-recycling rimless wheel: (a) Overall assembly, (b) Exploded view of the energy-recycling mechanism of one leg.}
 \label{CAD}
\end{figure}

The CAD overview of the energy-recycling rimless wheel is shown in Fig.~\ref{CAD}. The main body frame is a circular board. All eight legs are mounted on the main body, forming a rimless wheel. The energy-recycling mechanism of each leg consists of a spring, an electromagnetic clutch, a linear guide, and an optical flow sensor, as shown in Fig.~\ref{CAD}(b). The spring stores and releases elastic energy. The electromagnetic clutch locks and releases the leg relative to the main body during operation. When the electromagnetic clutch is powered on, the leg is locked to the board. When the power is off, the leg is free to move along the linear guide. The linear guide constrains the leg to move along the radial direction, prevents tangential displacement, and reduces friction during leg motion. An optical flow sensor is mounted on the board and monitors the motion of the leg, enabling timed clutch control. The foot is designed as a sharp inverted triangle to approximate the ideal contact point defined in the model. In addition, its wide profile helps suppress undesired motion outside the intended 2D plane. 

\subsection{Prototype and Electronic System Architecture}

\begin{figure}[H]
 \centering
 \includegraphics{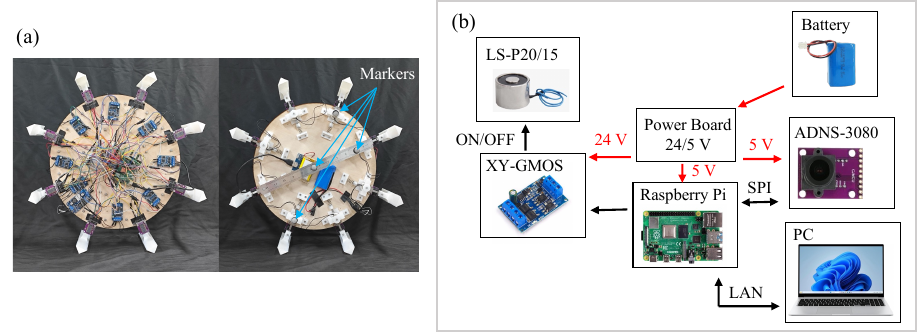}
 \caption{Physical prototype and electronic system architecture of the energy-recycling rimless wheel}
 \label{Propotype}
\end{figure}

A prototype of the proposed concept was built, as shown in Fig.~\ref{Propotype}(a). The main body frame was designed as a circular wooden board and fabricated using laser cutting. The leg structures were 3D printed using polylactic acid (PLA) for lightweight fabrication. The prototype includes electromagnetic clutches (model: LS-P20/15), XY-GMOS switching modules, an ADNS-3080 optical flow sensor, and a Raspberry Pi 4B. Four markers are placed in the same plane to enable optical tracking by the motion capture system.

As shown in Fig.~\ref{Propotype}(b), the electronic system is organized around the onboard Raspberry Pi 4B. The Raspberry Pi serves as the main controller, collecting leg-motion information from the ADNS-3080 optical flow sensor through the SPI interface and sending control commands to the XY-GMOS switching modules. The XY-GMOS modules control clutch engagement and disengagement by switching the clutch power supply. A PC is connected to the Raspberry Pi through a local area network (LAN) for remote operation and data monitoring.

\subsection{Working Principle}

\begin{figure}[H]
 \centering
 \includegraphics{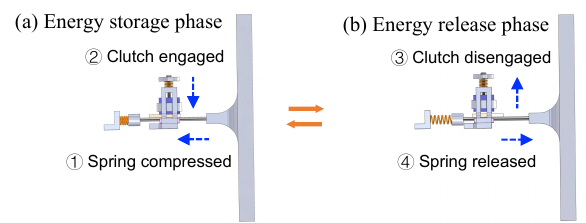}
 \caption{Working principle of the energy-recycling system: (a) Energy storage phase. The spring is compressed, and the clutch is engaged to lock the linear guide. (b) Energy release phase. The clutch is disengaged, and the stored spring energy is released forward.}
 \label{working principle}
\end{figure}

Figure~\ref{working principle} illustrates the operating process of the system. When the leg touches the ground, the spring is compressed; this leg is called the trailing leg. The leg that will touch the ground next is called the leading leg. The motion is monitored by the optical flow sensor. The maximum spring compression is detected when the measured leg velocity crosses zero. The Raspberry Pi collects displacement data from the sensor at a sampling rate of 300 Hz. The onboard program is designed to detect zero velocity and trigger a control signal to engage the clutch. In this way, energy is stored in the compressed spring in the trailing leg. When the robot's leading leg touches the ground, the optical flow sensor detects the leg motion. After this motion is detected, a command is sent to disengage the stance-leg clutch immediately so that the stored energy in the compressed spring is reinjected into the system. This process stores elastic energy in one step and releases it in the next step, supporting the intended energy-recycling mechanism.

\subsection{Experimental Validation}

\begin{figure}[H]
 \centering
 \includegraphics{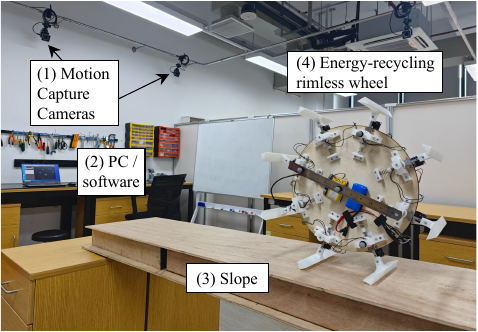}
 \caption{Experimental environment setup: (1) Motion capture cameras. (2) Vicon Tracker 3.8 motion capture software. (3) Walking slope. (4) Energy-recycling rimless wheel.}
 \label{setup}
\end{figure}

To examine the walking behavior of the prototype and compare it with the simulation, experiments are presented in this section. Figure~\ref{setup} shows the experimental setup layout. A Vicon motion capture system is used to track the motion of the energy-recycling rimless wheel. The system comprises eight cameras that record the position and orientation of the rigid body. Vicon Tracker 3.8 software is used for initial data processing and recording. Retro-reflective spherical markers are attached to the prototype, allowing the Vicon motion capture system to track its motion as it rolls down the constructed slope.

\begin{figure}[H]
  \centering
  \includegraphics{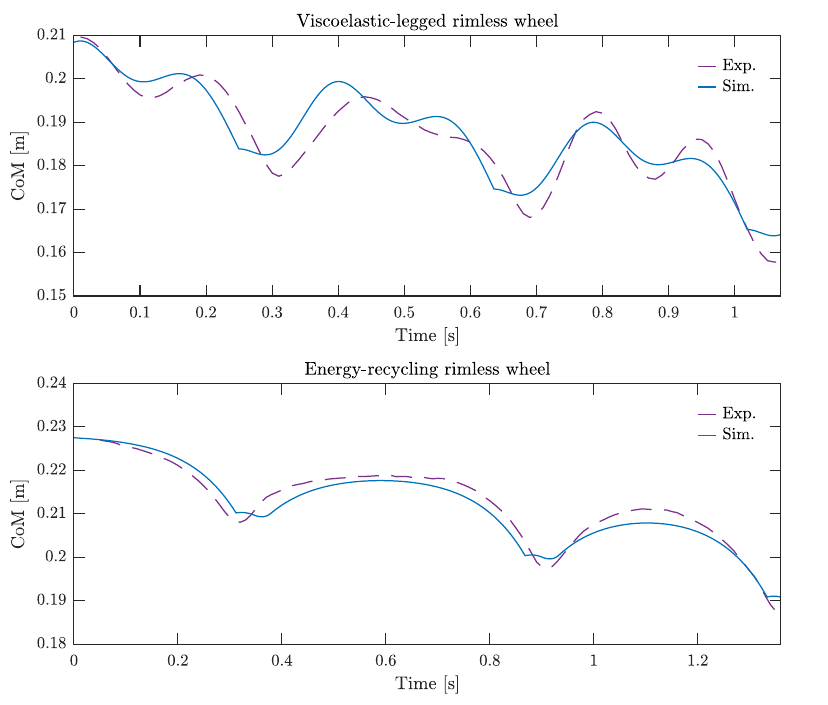}
  \caption{Comparison of experimental and simulated vertical CoM trajectories during passive downhill walking}
  \label{Z0_comparison}
\end{figure}

The mathematical model of the energy-recycling rimless wheel was numerically simulated for comparison with the experimental results, as shown in Fig.~\ref{Z0_comparison}. Exact agreement between the experimental and simulated trajectories is difficult to obtain because of several differences. A fully passive dynamic system is highly sensitive to initial conditions, which are difficult to measure accurately in experiments.

In addition, non-ideal material properties and unmodeled dissipative effects may affect the system response. These effects include material damping of the 3D-printed PLA components, friction in the linear guides, friction at the clutch-guide interface, additional resistance caused by slight assembly misalignment, and minor rolling or slipping effects at the foot--ground contact. Such losses can dissipate part of the released elastic energy and post-impact kinetic energy, and may therefore contribute to the discrepancy between the experimental and simulated trajectories.

A more detailed quantitative characterization of these effects can be obtained in future work through independent calibration experiments. For example, the effective damping of the compliant leg and the friction in the linear guide can be estimated from dedicated component-level tests. Incorporating these calibrated damping and friction parameters into the simulation model would further improve the agreement between simulation and experiment. Despite these differences, the experimental data and simulation results show qualitatively consistent trends, which support the feasibility of the proposed mechanism within the tested conditions.

\begin{figure}[H]
 \centering
 \includegraphics{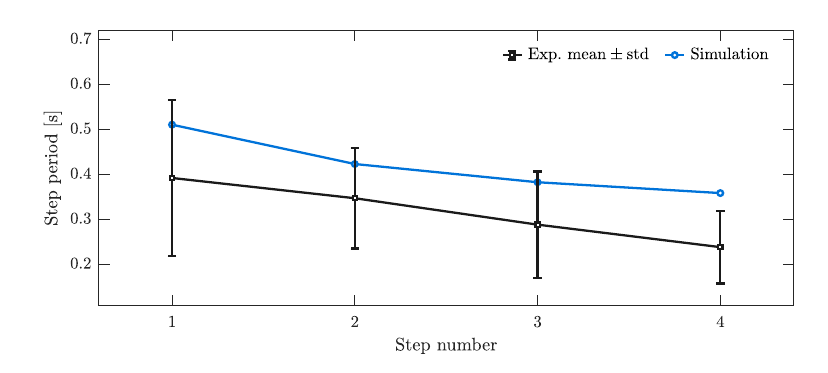}
 \caption{Comparison of step-period trends between the energy-recycling simulation and multiple prototype experiments on $1^\circ$--$3^\circ$ slopes}
 \label{fig:step_period_repeatability}
\end{figure}

Figure~\ref{fig:step_period_repeatability} compares the step-period trend obtained from the energy-recycling simulation with the experimental measurements of the prototype. The experimental points represent the mean value and standard deviation calculated from multiple passive walking trials conducted on slopes in the range of $1^\circ$--$3^\circ$. The measured step period decreases with the step number, and this tendency is also observed in the simulation. The experimental values are lower than the simulated values, especially in the later steps. This difference is consistent with the model--experiment mismatch discussed above and may also reflect the limited number of consecutive steps available on the current inclined platform. Therefore, the prototype exhibits a step-period trend that is qualitatively consistent with the simulation within the tested slope range, while a longer platform or treadmill-based setup is still required for a more systematic evaluation of long-term periodic behavior.

\begin{figure}[H]
 \centering
 \includegraphics{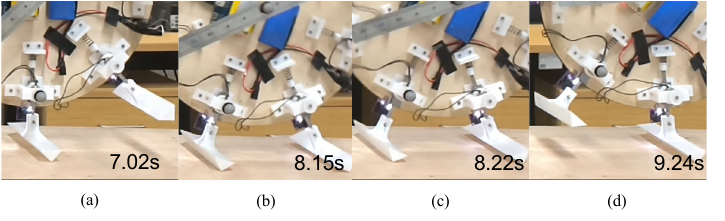}
 \caption{Snapshots of one passive walking cycle on a $1^\circ$ slope}
 \label{snapshots}
\end{figure}

Figure~\ref{snapshots} shows snapshots from the walking experiments. The spring-clutch workflow was implemented on the prototype. In the experiments, the slope angle was gradually decreased, and the robot was released until no smaller slope angle could sustain walking. This angle was defined as the critical angle. A critical angle of 1 degree was obtained from the experiment, indicating a CoT of 0.02. This value is of the same order as the CoT predicted by the simulation.

The present experiment does not fully validate the simulated $16.13\%$ reduction in CoT relative to the viscoelastic-legged configuration. That reduction is obtained from the numerical comparison reported in the numerical simulation section, where the minimum stable slope angle is used as the CoT metric. The prototype can in principle be tested with the clutch enabled and disabled, but a rigorous quantitative comparison in passive downhill walking would require more repeatable release conditions and a longer experimental platform. Therefore, the experimental results should be interpreted as preliminary validation of the spring-clutch mechanism and walking behavior, while the quantitative energy-efficiency improvement remains a simulation-based prediction.

\section{Conclusion and Discussion}
\label{sec:conclusion}
This paper proposed an energy-recycling rimless wheel with spring-clutch legs for passive downhill walking. By locking the compressed spring after foot contact and releasing the stored elastic energy in the subsequent step, the clutch separates the energy-storage process from the energy-release process while retaining the simple structure of the rimless wheel model. This mechanism provides a simple way to examine how clutch-regulated elastic energy return affects passive walking efficiency.

To analyze the proposed mechanism, we developed a hybrid dynamic model and compared it with a viscoelastic-legged rimless wheel. Periodic gaits were computed and their local stability was examined using a Poincaré return map. In the numerical comparison, the energy-recycling model reduced the CoT by up to $16.13\%$ compared with the viscoelastic-legged model within the tested stiffness range. The corresponding periodic gaits also remained locally stable in the numerical analysis.

The physical feasibility of the proposed design was examined using a prototype. The prototype achieved passive downhill walking on a $1^\circ$ slope, corresponding to a CoT of approximately $0.02$. The measured gait trends were qualitatively consistent with the simulation results. These results provide preliminary evidence for implementing the spring-clutch mechanism in a passive rimless wheel prototype, while the limitations of passive downhill walking experiments prevent a complete quantitative validation of the predicted CoT reduction at this stage.

The present analysis is limited by several modeling and experimental assumptions. The ideal rigid-contact formulation may introduce artificial contact-force transients during the double-support transition, and a more physically consistent treatment would require compliant-ground or penalty-based contact formulations. The experimental results are also affected by the available walking distance, release conditions, and surface friction. In addition, the electronic power consumption of the clutch and sensing system was not included in the present CoT evaluation and should be considered in future system-level efficiency studies.

The mechanism underlying the parameter-dependent energy efficiency also remains to be clarified. One possible explanation is that the system may approach resonance under certain parameter conditions, which could affect the energy transfer between spring compression, clutch locking, and forward motion. This effect will be investigated in future work.

An important next step is to extend the prototype toward level-ground walking with actuation. Such a platform would allow clutch-enabled and clutch-disabled conditions to be compared under the same mechanical configuration using measured actuator work or electrical input. Further work should include improved contact modeling and lower-energy locking mechanisms inspired by bistable or electroadhesive clutch designs \cite{plooij2015lock, krimsky2024elastic, diller2018effects}. Although energy-recycling mechanisms are not yet widely implemented in legged robots, the proposed approach may provide a useful direction for improving locomotion efficiency in more complex legged robotic systems.

\bibliography{\bib}

\begin{thebibliography}{10}
\providecommand{\url}[1]{#1}
\csname url@samestyle\endcsname
\providecommand{\newblock}{\relax}
\providecommand{\bibinfo}[2]{#2}
\providecommand{\BIBentrySTDinterwordspacing}{\spaceskip=0pt\relax}
\providecommand{\BIBentryALTinterwordstretchfactor}{4}
\providecommand{\BIBentryALTinterwordspacing}{\spaceskip=\fontdimen2\font plus
\BIBentryALTinterwordstretchfactor\fontdimen3\font minus \fontdimen4\font\relax}
\providecommand{\BIBforeignlanguage}[2]{{%
\expandafter\ifx\csname l@#1\endcsname\relax
\typeout{** WARNING: IEEEtran.bst: No hyphenation pattern has been}%
\typeout{** loaded for the language `#1'. Using the pattern for}%
\typeout{** the default language instead.}%
\else
\language=\csname l@#1\endcsname
\fi
#2}}
\providecommand{\BIBdecl}{\relax}
\BIBdecl

\bibitem{collins2005efficient}
S.~Collins, A.~Ruina, R.~Tedrake, and M.~Wisse, ``Efficient bipedal robots based on passive-dynamic walkers,'' \emph{Science}, vol. 307, no. 5712, pp. 1082--1085, 2005.

\bibitem{mcgeer1990passive}
T.~McGeer \emph{et~al.}, ``Passive dynamic walking,'' \emph{Int. J. Robotics Res.}, vol.~9, no.~2, pp. 62--82, 1990.

\bibitem{1570404}
S.~Collins and A.~Ruina, ``A bipedal walking robot with efficient and human-like gait,'' in \emph{Proceedings of the 2005 IEEE International Conference on Robotics and Automation}, 2005, pp. 1983--1988.

\bibitem{bhounsule2014low}
P.~A. Bhounsule, J.~Cortell, A.~Grewal, B.~Hendriksen, J.~D. Karssen, C.~Paul, and A.~Ruina, ``Low-bandwidth reflex-based control for lower power walking: 65 km on a single battery charge,'' \emph{The International Journal of Robotics Research}, vol.~33, no.~10, pp. 1305--1321, 2014.

\bibitem{7270326}
D.~Renjewski, A.~Spröwitz, A.~Peekema, M.~Jones, and J.~Hurst, ``Exciting engineered passive dynamics in a bipedal robot,'' \emph{IEEE Transactions on Robotics}, vol.~31, no.~5, pp. 1244--1251, 2015.

\bibitem{9099094}
L.~Li, I.~Tokuda, and F.~Asano, ``Energy-efficient locomotion generation and theoretical analysis of a quasi-passive dynamic walker,'' \emph{IEEE Robotics and Automation Letters}, vol.~5, no.~3, pp. 4305--4312, 2020.

\bibitem{coleman2010dynamics}
M.~J. Coleman, ``Dynamics and stability of a rimless spoked wheel: a simple 2d system with impacts,'' \emph{Dynamical Systems}, vol.~25, no.~2, pp. 215--238, 2010.

\bibitem{zheng2025tensegrity}
Y.~Zheng, F.~Asano, C.~Yan, L.~Li, and I.~T. Tokuda, ``Tensegrity-based legged robot generates passive walking, skipping, and crawling gaits in accordance with environment,'' \emph{IEEE/ASME Transactions on Mechatronics}, 2025.

\bibitem{bhounsule2016dead}
P.~A. Bhounsule, E.~Ameperosa, S.~Miller, K.~Seay, and R.~Ulep, ``Dead-beat control of walking for a torso-actuated rimless wheel using an event-based, discrete, linear controller,'' in \emph{International Design Engineering Technical Conferences and Computers and Information in Engineering Conference}, vol. 50152.\hskip 1em plus 0.5em minus 0.4em\relax American Society of Mechanical Engineers, 2016, p. V05AT07A042.

\bibitem{sanchez2021design}
S.~Sanchez and P.~A. Bhounsule, ``Design, modeling, and control of a differential drive rimless wheel that can move straight and turn,'' \emph{Automation}, vol.~2, no.~3, pp. 98--115, 2021.

\bibitem{hanazawa2018development}
Y.~Hanazawa, ``Development of rimless wheel with controlled wobbling mass,'' in \emph{2018 IEEE/RSJ International Conference on Intelligent Robots and Systems (IROS)}.\hskip 1em plus 0.5em minus 0.4em\relax IEEE, 2018, pp. 4333--4339.

\bibitem{gomes2011walking}
M.~Gomes and A.~Ruina, ``Walking model with no energy cost,'' \emph{Physical Review E—Statistical, Nonlinear, and Soft Matter Physics}, vol.~83, no.~3, p. 032901, 2011.

\bibitem{gomes2015quiet}
M.~W. Gomes and K.~Ahlin, ``Quiet (nearly collisionless) robotic walking,'' in \emph{2015 IEEE International Conference on Robotics and Automation (ICRA)}.\hskip 1em plus 0.5em minus 0.4em\relax IEEE, 2015, pp. 5761--5766.

\bibitem{alexander1990three}
R.~Alexander \emph{et~al.}, ``Three uses for springs in legged locomotion,'' \emph{International Journal of Robotics Research}, vol.~9, no.~2, pp. 53--61, 1990.

\bibitem{pal2026energy}
P.~Pal, S.~Kolathaya, and A.~Ghosal, ``Energy-efficient quadruped locomotion with compliant feet,'' \emph{arXiv preprint arXiv:2605.14411}, 2026.

\bibitem{rummel2010stable}
J.~Rummel, Y.~Blum, H.~M. Maus, C.~Rode, and A.~Seyfarth, ``Stable and robust walking with compliant legs,'' in \emph{2010 IEEE International Conference on Robotics and Automation}.\hskip 1em plus 0.5em minus 0.4em\relax IEEE, 2010, pp. 5250--5255.

\bibitem{hurst2008role}
J.~W. Hurst, \emph{The role and implementation of compliance in legged locomotion}.\hskip 1em plus 0.5em minus 0.4em\relax Carnegie Mellon University, 2008.

\bibitem{collins2015reducing}
S.~H. Collins, M.~B. Wiggin, and G.~S. Sawicki, ``Reducing the energy cost of human walking using an unpowered exoskeleton,'' \emph{Nature}, vol. 522, no. 7555, pp. 212--215, 2015.

\bibitem{collins2010recycling}
S.~H. Collins and A.~D. Kuo, ``Recycling energy to restore impaired ankle function during human walking,'' \emph{PLoS one}, vol.~5, no.~2, p. e9307, 2010.

\bibitem{wang2021design}
C.~Wang, L.~Dai, D.~Shen, J.~Wu, X.~Wang, M.~Tian, Y.~Shi, and C.~Su, ``Design of an ankle exoskeleton that recycles energy to assist propulsion during human walking,'' \emph{IEEE Transactions on Biomedical Engineering}, vol.~69, no.~3, pp. 1212--1224, 2021.

\bibitem{lee2023lower}
H.~Lee and J.~Rosen, ``Lower limb exoskeleton-energy optimization of bipedal walking with energy recycling-modeling and simulation,'' \emph{IEEE Robotics and Automation Letters}, vol.~8, no.~3, pp. 1579--1586, 2023.

\bibitem{krimsky2024elastic}
E.~Krimsky and S.~H. Collins, ``Elastic energy-recycling actuators for efficient robots,'' \emph{Science Robotics}, vol.~9, no.~88, p. eadj7246, 2024.

\bibitem{badri2022birdbot}
A.~Badri-Spr{\"o}witz, A.~Aghamaleki~Sarvestani, M.~Sitti, and M.~A. Daley, ``Birdbot achieves energy-efficient gait with minimal control using avian-inspired leg clutching,'' \emph{Science Robotics}, vol.~7, no.~64, p. eabg4055, 2022.

\bibitem{asano2012passive}
F.~Asano and J.~Kawamoto, ``Passive dynamic walking of viscoelastic-legged rimless wheel,'' in \emph{2012 IEEE International Conference on Robotics and Automation}.\hskip 1em plus 0.5em minus 0.4em\relax IEEE, 2012, pp. 2331--2336.

\bibitem{gamus2015dynamic}
B.~Gamus and Y.~Or, ``Dynamic bipedal walking under stick-slip transitions,'' \emph{SIAM Journal on Applied Dynamical Systems}, vol.~14, no.~2, pp. 609--642, 2015.

\bibitem{plooij2015lock}
M.~Plooij, G.~Mathijssen, P.~Cherelle, D.~Lefeber, and B.~Vanderborght, ``Lock your robot: A review of locking devices in robotics,'' \emph{IEEE Robotics \& Automation Magazine}, vol.~22, no.~1, pp. 106--117, 2015.

\bibitem{diller2018effects}
S.~B. Diller, S.~H. Collins, and C.~Majidi, ``The effects of electroadhesive clutch design parameters on performance characteristics,'' \emph{Journal of Intelligent Material Systems and Structures}, vol.~29, no.~19, pp. 3804--3828, 2018.

\end{thebibliography}

\end{document}